\documentclass[runningheads]{llncs}
\usepackage{graphicx}
\usepackage{subfig}
\usepackage{amsmath,amssymb} 
\usepackage{color}
\usepackage{booktabs}
\usepackage{enumerate}
\usepackage[title]{appendix}
\usepackage[width=122mm,left=12mm,paperwidth=146mm,height=193mm,top=12mm,paperheight=217mm]{geometry}
\usepackage[pagebackref=true,breaklinks=true,letterpaper=true,colorlinks,bookmarks=false]{hyperref}
\begin{document}
	\def\ECCV18SubNumber{1232}  
	
	\title{An Improved Evaluation Framework for Generative Adversarial Networks} 
		
	\author{Shaohui Liu$^*$, Yi Wei$^*$, Jiwen Lu, Jie Zhou}
	\institute{Department of Automation\\Tsinghua University\\\{\texttt{sh-liu15, wei-y15}\}\texttt{@mails.tsinghua.edu.cn}\\ \{\texttt{lujiwen, jzhou}\}\texttt{@tsinghua.edu.cn}}
	
	\maketitle
	\renewcommand{\thefootnote}{\fnsymbol{footnote}}
	\begin{abstract}
		\footnotetext[1]{equal contribution}
		In this paper, we propose an improved quantitative evaluation framework for Generative Adversarial Networks (GANs) on generating domain-specific images, where we improve conventional evaluation methods on two levels: the feature representation and the evaluation metric. Unlike most existing evaluation frameworks which transfer the representation of ImageNet inception model to map images onto the feature space, our framework uses a specialized encoder to acquire fine-grained domain-specific representation. Moreover, for datasets with multiple classes, we propose Class-Aware Frechet Distance (CAFD), which employs a Gaussian mixture model on the feature space to better fit the multi-manifold feature distribution. Experiments and analysis on both the feature level and the image level were conducted to demonstrate improvements of our proposed framework over the recently proposed state-of-the-art FID method. To our best knowledge, we are the first to provide counter examples where FID gives inconsistent results with human judgments. It is shown in the experiments that our framework is able to overcome the shortness of FID and improves robustness. Code will be made available\footnote[4]{\url{https://github.com/B1ueber2y/CAFD}}.  
		
	\end{abstract}
	
	\section{Introduction}
	
	Generative Adversarial Networks (GANs) have shown outstanding abilities on many computer vision tasks including generating domain-specific images \cite{NIPS2014_5423}, style transfer \cite{Zhu_2017_ICCV}, super resolution \cite{Ledig_2017_CVPR}, etc. The basic idea of GANs is to hold a two-player game between generator and discriminator, where the discriminator aims to distinguish between real and fake samples while the generator tries to generate samples as real as possible to fool the discriminator. 
	
	Researchers \cite{DBLP:journals/corr/RadfordMC15,arjovsky2017wasserstein,DBLP:journals/corr/BerthelotSM17,Mao_2017_ICCV} have been continuously exploring better GAN architectures. However, developing a widely-accepted GAN evaluation framework remains to be a challenging topic \cite{Theis2016a}. Due to lack of GAN benchmark results, newly proposed GAN variants are validated on different evaluation frameworks and therefore incomparable. Because human judgements are inherently limited by manpower resource, good quantitative evaluation frameworks are of very high importance to guide future research on designing, selecting, and interpreting GAN models.  
	
	\begin{figure}[tb]
		\centering
		\includegraphics[width=4.5in]{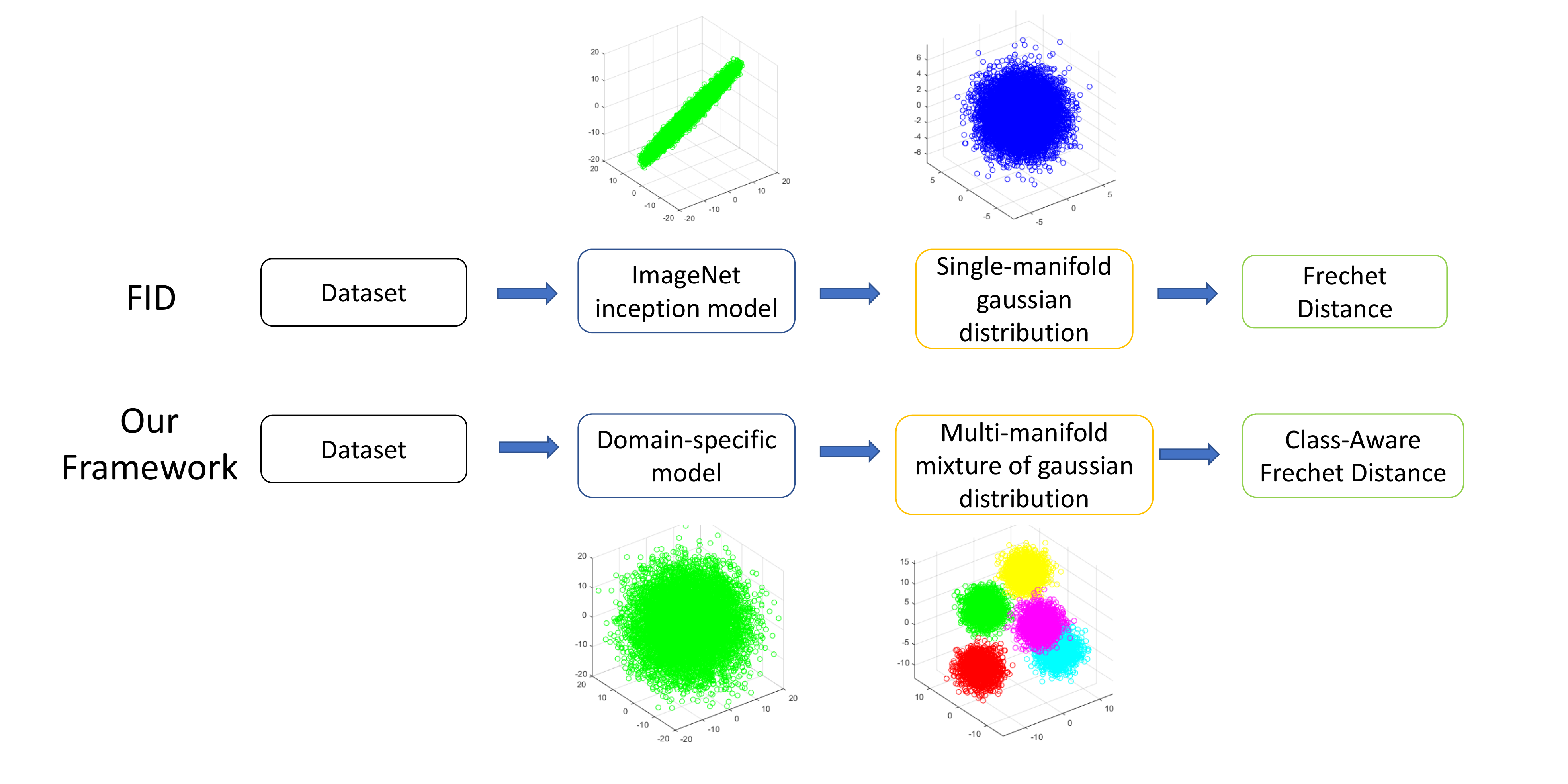}
		\caption{Comparison between our proposed framework and the recently proposed state-of-the-art evaluation method FID \cite{NIPS2017_7240}. Our framework uses a domain-specific representation to get better features and employs a multi-manifold Gaussian mixture model to better fit the distribution.}
		\label{fig::pipeline}
	\end{figure}
	
	There have been varieties of efforts on evaluating GANs on its ability of generating domain-specifc images. The goal is to measure the distance between the generated samples and the real in the dataset. Most existing methods utilized the ImageNet \cite{ILSVRC15} inception model to map images onto the feature space. The most widely used criteria is probably the Inception Score \cite{NIPS2016_6125}, which measures the distance via Kullback-Leiber Divergence (KLD). However, it is probability based and is unable to report overfitting. Recently, Frechet Inception Distance (FID) was proposed \cite{NIPS2017_7240} on improving Inception Score. It directly measures Frechet Distance on the feature space with the single-manifold Gaussian assumption. It has been proved that FID is far better than Inception Score \cite{huang2018an,jiwoong2018quantitatively,2017arXiv171110337L}. However, we argue that assuming normality on the whole feature distribution may lose class information on labeled datasets. 
	
	In this work, we propose an improved quantitative evaluation framework. Comparison between our framework and the current state-of-the-art FID method is shown in Fig. \ref{fig::pipeline}. We improve conventional evaluation methods on two levels: the feature representation and the evaluation metric. Unlike most existing methods including the Inception Score \cite{NIPS2016_6125} and FID \cite{NIPS2017_7240}, our framework uses a specialized encoder trained on the dataset to get domain-specific representation. We argue that applying the ImageNet model to either labeled or unlabeled datasets is ineffective. Moreover, we propose Class-Aware Frechet Distance (CAFD) in our framework to measure the distribution distance of each class (mode) respectively on the feature space to include class information. Instead of the single-manifold Gaussian assumption, we employ a Gaussian mixture model (GMM) to better fit the feature distribtution. We also include KL divergence (KLD) between mode distribution of real data and generated samples into the framework to help detect mode dropping. 
	
	Experiments and analysis on both the feature level and the image level were conducted to demonstrate the improved effectiveness of our proposed framework. To our best knowledge, we are the first \cite{2018arXiv180203446B} to provide counter examples where FID is inconsistent with human judgements (See Figs. \ref{fig::hackimagenet} and \ref{fig::hackfid}). It is shown in the experiments that our framework is able to overcome the shortness of existing methods. 
	
	\section{Related Work}
	
	\noindent
	\textbf{Generative Adversarial Networks.}
	The idea of Generative Adversarial Network was originally proposed in \cite{NIPS2014_5423}. It has been applied to various computer vision tasks \cite{Zhu_2017_ICCV,Ledig_2017_CVPR,zhu2016generative,Isola_2017_CVPR}. Researchers have been continuously developing better GAN architectures \cite{Gurumurthy_2017_CVPR,Huang_2017_CVPR} and training strategies \cite{DBLP:conf/icml/Arora0LMZ17,hoang2018mgan} on generating domain-specific images. Deep convolutional networks were firstly introduced to the GAN community by \cite{DBLP:journals/corr/RadfordMC15}. Wasserstein GAN (WGAN) \cite{arjovsky2017wasserstein} was proposed to significantly improve convergence on GAN training. Recently, several variants were proposed \cite{DBLP:journals/corr/BerthelotSM17,Mao_2017_ICCV,che2015mode,dziugaite2015training,zhao2016energy,DBLP:journals/corr/KodaliAHK17,NIPS2017_7159} to improve the image quality generated by GAN models. 
	
	~\\
	\noindent
	\textbf{Evaluation Methods.}
	Several GAN evaluation methods have been proposed by researchers. While model-based methods including Parzen window estimation and the annealed importance sampling (AIS) \cite{wu2016quantitative} require either density estimation or observation on the inner structure of the decoder, model-agnostic methods \cite{NIPS2017_7240,NIPS2016_6125,jiwoong2018quantitatively,che2015mode,dziugaite2015training,lopez2016revisiting,li2015generative} are more popular in the GAN community. These methods are sample based. Most of them map images onto the feature space via an ImageNet pretrained model and measure the similarity of the distribution between the dataset and the generated data. Maximum mean discrepancy (MMD) was proposed in \cite{dziugaite2015training,li2015generative} and it has been further used in classifier two-sample tests \cite{lopez2016revisiting}, where statistical hypothesis testing is used to assess whether two sample sets are from the same distribution. Inception Score \cite{NIPS2016_6125}, along with its improved version Mode Score \cite{che2015mode}, was the most widely used metric in the last two years. Recently, FID \cite{NIPS2017_7240} was proposed on improving the Inception Score.
	
	~\\
	\noindent
	\textbf{Studies on Existing Frameworks.}
	It is common \cite{2018arXiv180101973B} in the literature to see algorithms which use existing metrics to optimize early stopping, hyperparameter tuning, and even model architecture. Thus, comparison and analysis on previous evaluation methods have been attracting more and more attention recently \cite{Theis2016a,huang2018an,jiwoong2018quantitatively,2017arXiv171110337L}. While Inception Score was the most popular metric in the last two years, it was believed to be misleading in recent literature \cite{NIPS2017_7240,huang2018an,2017arXiv171110337L,2018arXiv180203446B,2018arXiv180101973B}. Applying the ImageNet model to encode features in Inception Score is ineffective \cite{Theis2016a,2018arXiv180101973B,2017arXiv170604987R}. The recently proposed FID has been proved to be far better than Inception Score \cite{NIPS2017_7240,huang2018an,jiwoong2018quantitatively}. And its robustness was experimentally demonstrated recently in a technical report by Google Brain \cite{2017arXiv171110337L}. However, in this paper, we argue that FID still has problems and provide counter examples where FID gives inconsistent results with human judgements. Moreover, we propose an improved evaluation framework and overcome the shortness of existing methods.
	
	\section{Problems on FID}
	\subsection{Method Formulation}
	The evaluation problem can be formulated as modeling the distance between two distributions $P_r$ and $P_g$, where $P_r$ denotes the distribution of real samples in the dataset and $P_g$ denotes the distributions of new samples generated by GAN models.
	
	The main difficulties for GANs on generating domain-specific images can be summarized into three types below.
	
	\begin{itemize}	
		\item[$\bullet$] \textbf{Lack of generating ability.} Either the generator cannot generate useful samples or the GAN training cannot diverge. 
		
		\item[$\bullet$] \textbf{Mode collapse.} Different modes collapse to a new mixed mode in the generated samples. (e.g. An animal resembling both a horse and a deer.)
		
		\item[$\bullet$] \textbf{Mode dropping}. Only part of the modes in the dataset are generated while some modes are implicitly ignored. (e.g. The handwritten ‘5’ can hardly be generated by GAN trained on MNIST.) 
	\end{itemize}
	
	Therefore, a good evaluation framework should be consistent to human judgements, penalize on mode collapse and mode dropping. 
	
	Most of the conventional methods utilized an ImageNet pretrained inception model to map images onto the feature space. Inception Score, which was originally formulated as Eq. (\ref{eq::is}), ignored information in the dataset completely. Thus, its original formulation was considered to be relatively misleading.
	\begin{equation}
		\label{eq::is}
		IS=exp(E_x[KL(p(y|x)||p(y))])
	\end{equation}
	
	The Mode Score was proposed \cite{che2015mode} to overcome this shortness. Its formulation is shown in Eq. (\ref{eq::ms}). By including the prior distribution of the ground truth labels, Mode Score improved Inception Score \cite{che2015mode} on reporting mode dropping. 
	
	\begin{equation}
		\label{eq::ms}
		MS=exp(E_x[KL(p(y|x)||p(y^*))]-KL(p(y^*)||p(y)))
	\end{equation}
	
	FID \cite{NIPS2017_7240}, which was formulated in Eq. (\ref{eq::fid}), was proposed on improving Inception Score \cite{NIPS2016_6125}. Unlike the previous two metrics which are probability-based, FID directly measures Frechet distance on the feature space. It assumes single-manifold normality on the feature distribution and uses an ImageNet model for encoding features. FID was believed to be better than Inception Score \cite{huang2018an,jiwoong2018quantitatively,2017arXiv171110337L}. However, we argue that FID has two major problems (See Section \ref{sec::fid_rep} and \ref{sec::fid_multim}). 
	
	\begin{equation}
		\label{eq::fid}
		FID(P_r, P_g)=||\mu_r-\mu_g||+Tr(C_r+C_g-2(C_rC_g)^\frac{1}{2})
	\end{equation}

	\subsection{Ineffective Representation}
	\label{sec::fid_rep}
	As both Inception Score \cite{NIPS2016_6125} and Mode Score \cite{che2015mode} is probability-based, applying the ImageNet pretrained model on non-ImageNet dataset is relatively meaningless. This misuse of representation on Inception Score was mentioned previously \cite{2017arXiv170604987R}. However, we argue that applying the ImageNet model to map the generated images to the feature space in FID can also be misleading. 
	
	On labeled dataset with multiple classes, the class labels unmatch those in ImageNet. For example, the class `Bird' in CIFAR-10 \cite{krizhevsky2009learning} is divided into several sophisticated category labels in ImageNet. Therefore, the CNN representation trained on ImageNet is either meaningless or over-complicated.
	
	On unlabeled dataset with images from a single class such as CelebA \cite{liu2015faceattributes} and LSUN Bedrooms \cite{yu15lsun}, applying the ImageNet inception model is also inapproriate. The categories of ImageNet labels are so sophisticated that the trained model needs to encode diverse features on various objects. However, the learned features are ineffective on a specific domain. The encoded features are limited to a relatively low-dimensional manifold lack of fine-grained information. In Section \ref{sec::hackimagenet}, we designed experiments on both the feature level and the image level to demonstrate the effects of representation. To our best knowledge, we are the first \cite{2018arXiv180203446B} to provide examples where FID gives misleading results on unlabeled datasets (See Fig. \ref{fig::hackimagenet}).
	
	\subsection{Single-Manifold vs. Multi-Manifold}
	\label{sec::fid_multim}
	
	We argue that the single-manifold multivariate Gaussian assumption in FID is considered to be over-simplified. As the training decreases intra-class distance and increases inter-class distance, the features are distributed in groups by their class labels. Thus, on datasets with multiple classes, the feature distribution is more like a multi-manifold structure, which is better fitted by a multivariate Gaussian mixture model (GMM).
	
	Considering the specific Gaussian mixture model where $x \sim N(\mu_i, C_i)$ with probability $p_i$, we can derive the first and second moment of the feature distribution in Eq. (\ref{eq::gmm_mu}) and Eq. (\ref{eq::gmm_C}).
	
	\begin{equation}
		\label{eq::gmm_mu}
		\mu=E(x)=E(E(x|y))=\sum{p_i\mu_i}
	\end{equation}
	\begin{equation}
		\label{eq::gmm_C}
		\begin{aligned}
			C&=\text{var}(x)=E(\text{var}(x|y))+\text{var}(E(x|y)) \\
			&=\sum{p_iC_i}+\sum{p_i(\mu_i-\mu)(\mu_i-\mu)^T}
		\end{aligned}
	\end{equation}
	
	It should be noted that when the feature is n-dimensional and there are $K$ classes in total, there are a total of $K(\frac{n^2+n}{2}+n+1)$ variables in the model. However, directly modeling the whole distribution Gaussian as in FID will result in $\frac{n^2+n}{2}+n$ degrees of freedom, which is a relatively small number. Thus, FID detects mode-related problems in a much implicit way. Although FID was proved to be robust to mode dropping and mode collapse in recent literature \cite{NIPS2017_7240,huang2018an,jiwoong2018quantitatively,2017arXiv171110337L}, we argue that experimental demonstrations on its robustness in previous work is insufficient. Either simply dropping a mode or linearly combining images will result in increased FID. However, in cases where the mode-related problems are more complicated, FID may give misleading results (See Fig. \ref{fig::hackfid}). In Section \ref{sec::exp_metric}, we conducted sufficient experiments to analyze the property of encoded features. To our best knowledge, we are the first to provide counter examples where FID fails to give consistent results with human judgements on datasets with multiple classes.

	\section{Proposed Framework}
	
	\subsection{Domain-Specific Encoder}
\begin{figure}[tb]
		\centering
		\begin{tabular}{cc}
			\subfloat[Representation]{\includegraphics[width=2.0in]{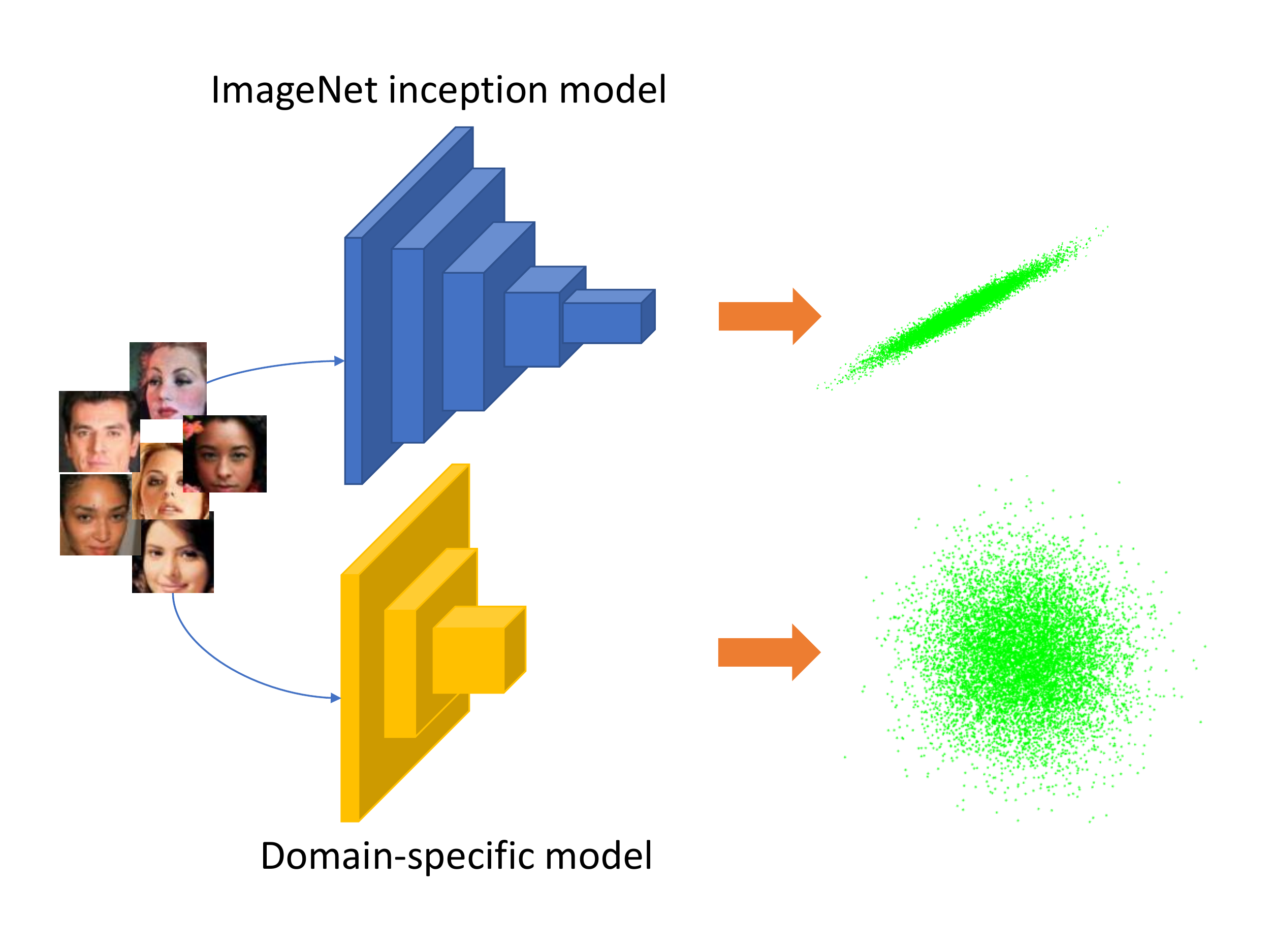}} & 
			\subfloat[Evaluation metric]{\includegraphics[width=2.0in]{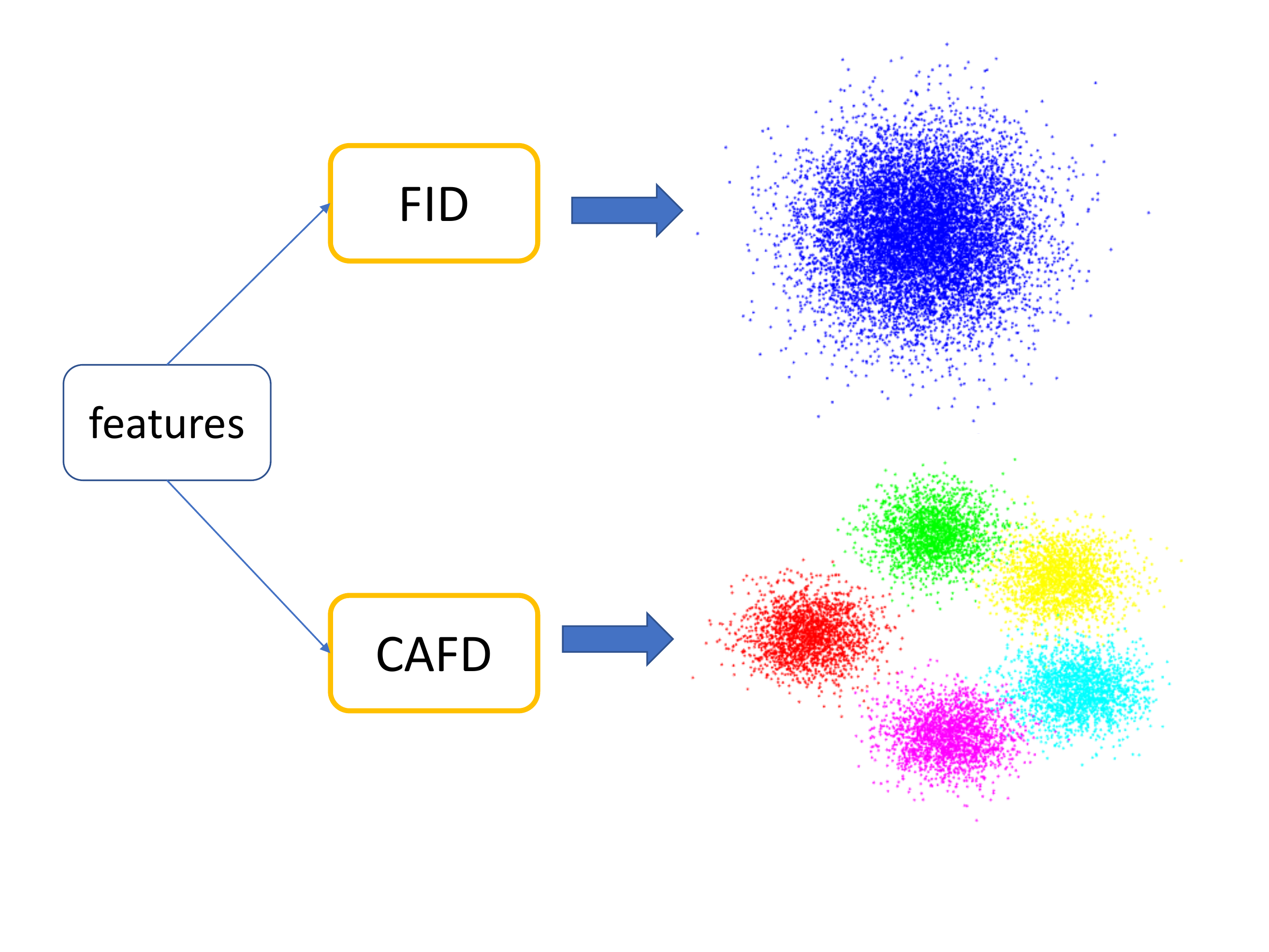}}
		\end{tabular}
		\caption{Visual demonstrations on highlights of our proposed framework. In the left figure, the features encoded by the ImageNet model are limited within a low-dimensional subspace. Thus, we propose that a domain-specific encoder is needed. In the right figure, we show that instead of a single-manifold Gaussian distribution, the features are more like a multi-manifold structure. CAFD employs a Gaussian mixture model to include class information.}
		\label{fig::motivations}
	\end{figure}

	As discussed in Section \ref{sec::fid_rep}, applying the ImageNet inception model to either labeled or unlabeled datasets is ineffective. Thus, we argue that a specialized domain-specific encoder should be used in the evaluation framework. As shown in Fig. \ref{fig::motivations}(a), while the features encoded by the ImageNet model are limited within a low-dimensional subspace, the domain-specific model could encode more fine-grained information, making the encoded features much more effective.
	
	Specifically, for datasets with multiple classes such as CIFAR-10 \cite{krizhevsky2009learning}, representation is acquired via training a domain-specific classifier. On dataset without class labels such as CelebA \cite{liu2015faceattributes},  the unsupervised representation learning method, specifically, AutoEncoder, is used to get more effective representation.

	\label{sec::sneeded}
	\subsection{Class-Aware Frechet Distance}
	Before introducing our improved evaluation metric, we would firstly take a step back towards existing popular metrics. Both Inception Score \cite{NIPS2016_6125} and Mode Score \cite{che2015mode} measure distance between probability distribution while FID \cite{NIPS2017_7240} directly measures distance on the feature space. These are two different perspectives towards evaluating GAN models. Probability-based metrics better handle mode-related problems (with the correct use of a domain-specific encoder), while directly measuring distance between features better models the generating ability. In fact, we believe these two perspectives are complementary. In our framework, we propose a class-aware metric on the feature space to combine the two perspectives together. 
	
	As shown in Fig. \ref{fig::motivations}(b), for datasets with multiple classes, the feature distribution is more like a multi-manifold structure (See Section \ref{sec::fid_multim}). Thus, we use a Gaussian mixture model (GMM) and propose Class-Aware Frechet Distance (CAFD) to include class information. Specifically, we compute probability-based Frechet Distance between real data and generated samples in each class respectively.
	
	As previously discussed in Section \ref{sec::sneeded}, we train a domain-specific classifier on datasets with multiple classes and use its derived representation. In our proposed framework, we also made use of the predicted probability $p(y|x)$. To calculate the expected mean of each class in a specific set $S$ of generated samples, we can derive the formulation below in Eq. (\ref{eq::cafd_mu}). 
	
	\begin{equation}
		\label{eq::cafd_mu}
		\begin{aligned}
			\mu^g_i &=E[x|y_i] = \sum_{x_j \in S}{x_j\text{p}(x_j|y_i)}=\sum_{x_j \in S}{x_j\frac{\text{p}(x_j,y_i)}{\text{p}(y_i)}} \\
			& = \sum_{x_j \in S}{x_j\frac{\text{p}(y_i|x_j)\text{p}(x_j)}{\sum_{x^* \in S}\text{p}(y_i|x^*)\text{p}(x^*)}} \\
			& \overset{i.i.d}{=}\sum_{x_j \in S}{x_j\frac{\text{p}(y_i|x_j)}{\sum_{x^* \in S}\text{p}(y_i|x^*)}}=\sum_{x_j \in S}{w_{ij}x_j}
		\end{aligned}
	\end{equation}
	
	\noindent
	where
	\begin{equation}
		w_{ij}=\frac{\text{p}(y_i|x_j)}{\sum_{x^* \in S}\text{p}(y_i|x^*)}
	\end{equation}
	
	Similarly, The covariance matrix in each class is shown in Eq. (\ref{eq::cafd_C}).
	\begin{equation}
		\label{eq::cafd_C}
		C^g_i=\sum_{x \in S}{w_{ij}(x_j-\mu_i)(x_j-\mu_i)^T}
	\end{equation}
	We compute Frechet distance in each of the $K$ classes and average the results to get Class-Aware Frechet Distance (CAFD) in Eq. (\ref{eq::cafd}).
	\begin{equation}
		\label{eq::cafd}
		CAFD(P_r, P_g)=\frac{1}{K}\sum_{i=1}^K{\{||\mu^r_i-\mu^g_i||+Tr(C^r_i+C^g_i-2(C^r_iC^g_i)^\frac{1}{2})\}}
	\end{equation}
	
	This improved form based on Gaussian mixture model assumption can better evaluate the actual distance than the original FID. Moreover, more comprehensive evaluation results can be derived. When CAFD is applied to evaluating a specific GAN model, we could get better class-aware understanding towards the generating ability. For example, as shown in Table \ref{tab::class}, the selected model generates digit 1 well but struggles on other classes. This information will provide guidance for researchers on how well their generative models perform on each mode and may explain what specific problems exist.
	
	\begin{table}[tb]
		\centering
		\setlength{\abovecaptionskip}{5pt}
		\caption{Frechet distance on different classes of MNIST dataset.}
		\begin{tabular}{*{7}{|c}|}
			\hline
			Class & 0 & 1 & 2 & 3 & 4 & 5\\
			\hline
			Distance & $210.6 \pm 5.5$ & $78.0 \pm 4.4$ & $299.0 \pm 14.5$ & $204.9 \pm 4.0$ & $218.3 \pm 6.6$ & $241.9 \pm 5.0$ \\
			\hline
			Class & 6 & 7 & 8 & 9 & \multicolumn{2}{c|}{average} \\
			\hline
			Distance & $207.8 \pm 3.2$ & $157.6 \pm 3.9$ & $212.8 \pm 3.2$ & $179.3 \pm 1.9$ & \multicolumn{2}{c|}{$201.0 \pm 0.9$} \\
			\hline
		\end{tabular}
		\label{tab::class}
	\end{table}
	
	As both FID and CAFD aim to model how well domain-specific images are generated, they are not designed to deal with mode dropping, where some of the modes are missed in the generated samples. We propose that both metrics detect mode dropping in a relatively implicit way, which may fail in some corner cases. Thus, motivated by Mode Score \cite{che2015mode}, we propose that KL divergence $KL(p(y^*)||p(y))$ should be included into the evaluation framework. 
	
	To sum up, the correct use of encoder, the CAFD and the KL divergence term combine for an complete improved evaluation framework. Our proposed method combines the advantages of Inception Score \cite{NIPS2016_6125}, Mode Score \cite{che2015mode} and FID \cite{NIPS2017_7240} and overcomes their shortness.
	
	\subsection{Discussion}
	Our method is sensitive to different representations. Different selection of encoders may result in changes on the evaluation results. Experiments in Section \ref{sec::exp_rep} demonstrate that the ImageNet inception model may give misleading results (See Fig. \ref{fig::hackimagenet}). Thus, a domain-specific encoder should be used in each evaluation framework. We argue that because the representation is not fixed, the correct use (with domain-specific representation) of Inception Score, Mode Score and FID would suffer from this sensitivity problem as well. It is worth emphasizing that different generative methods should be compared only under the same encoder.  
	
	Unlike Inception Score, because CAFD measures distance on the feature space as FID does, it is able to report overfitting. By measuring CAFD with respect to training set and test set respectively, researchers can get understanding towards whether their GAN models overfit the training data.  Moreover, the intermediate results could provide researchers comprehensive understanding towards their GAN models (e.g. See Table \ref{tab::class}). 
	
	\section{Experiments}
	
	\subsection{Study on Representation}
	\label{sec::exp_rep}
	In this section, we study the representation for mapping the generated images onto the feature space. As discussed in Section \ref{sec::sneeded}, applying the pretrained ImageNet inception model to either labeled or unlabeled datasets is considered to be inappropriate. We first investigated the problem of unmatched class labels on a labeled dataset, specifically, CIFAR-10 \cite{krizhevsky2009learning}. Then, experiments on both the feature space and image level were conducted on CelebA \cite{liu2015faceattributes}, which is a dataset including only face images. 
	
	\subsubsection{Experiments on CIFAR-10 \cite{krizhevsky2009learning}:}
	We used Inception-v3 \cite{Szegedy_2015_CVPR} model trained on ImageNet to classify the 5000 images labeled `Bird' and 5000 images labeled `Dog' in CIFAR-10 \cite{krizhevsky2009learning} dataset respectively. Table \ref{tab::cifar10onimagenet} shows the results. The images from the single class `Bird' in CIFAR-10 is classified into various subclasses, where surprisingly the top class is Fox Squirrel (which is not a Bird class) with a 10.1\% frequency. The classification results are extremely diverse. It can be inferred that the Inception-v3 model trained on ImageNet does not map images with the label `Bird' onto a single-manifold subspace. Results on the label `Dog' show similar patterns. We argue that features determining whether a dog is a Japanese spaniel or an English foxhound are unnecessary on CIFAR-10. The ImageNet representation cannot well fit non-ImageNet datasets.
	
	\begin{table}[tb]
		\centering
		\setlength{\abovecaptionskip}{5pt}
		\caption{The classification results on CIFAR-10 \cite{krizhevsky2009learning} images using inception model trained on ImageNet. The class labels 'Bird' and 'Dog' are divided into several subclasses.}
		\label{tab::cifar10onimagenet}
		\begin{tabular}{|p{1.0cm}|p{3.0cm}|p{1.5cm}||p{3.0cm}|p{1.5cm}|}
			\hline
			Rank & CIFAR-10 'Bird' & Frequency & CIFAR-10 'Dog' & Frequency \\
			\hline
			1 & Fox Squirrel & 10.1\% & Japanese spaniel & 9.8\% \\
			\hline
			2 & Limpkin & 6.9\% & Dandie Dinmont & 5.2\% \\
			\hline
			3 & Black Stork & 6.4\% & English foxhound & 4.6\% \\
			\hline
			4 & Black Grouse & 5.3\% & Toy terrier & 3.2\% \\
			\hline
			5 & Brambling & 4.1\% & Bluetick & 2.8\% \\
			\hline
		\end{tabular}
	\end{table}
	
	Therefore, when the dataset includes multiple classes and its class labels are different from those of ImageNet, the feature encoder should be specifically trained. To attain effective representation on non-ImageNet datasets, we need to ensure that the class labels of data used for training GAN models are consistent with those of data used for training the encoder. 
	
	\subsubsection{Experiments on CelebA \cite{liu2015faceattributes}:}
	Regardless of the unmatched classes problem, applying the ImageNet pretrained model to label-free dataset for GAN training can still give misleading results. Take the face dataset CelebA \cite{liu2015faceattributes} for example. On one hand, in order to evaluate how well the face images were generated, the encoder needs to encode facial texture features, which are hardly learned in the ImageNet inception model. On the other hand, the features determining whether a bird is a limpkin or a grouse are obviously unnecessary on CelebA. Thus, the percentage of effective features on the whole feature space is relatively low. 
	
	Experiments were conducted on the CelebA \cite{liu2015faceattributes} dataset to better demonstrate the deficiency of the ImageNet model. We performed three different types of adjustments on the first 10,000 images on CelebA: a) Random noise uniformly distributed in [-33,33] was applied on each pixel. b) Each image was divided into 8x8=64 regions and seven of them were sheltered by a pixel sampled from the face. c) Each image was first divided into 4x4=16 regions and random exchanges were performed twice. 
	
\begin{figure}[tb]
		\centering
		\begin{tabular}{ccccc}
			\subfloat[noise (FID=75.9)]{\includegraphics[width=1.5in]{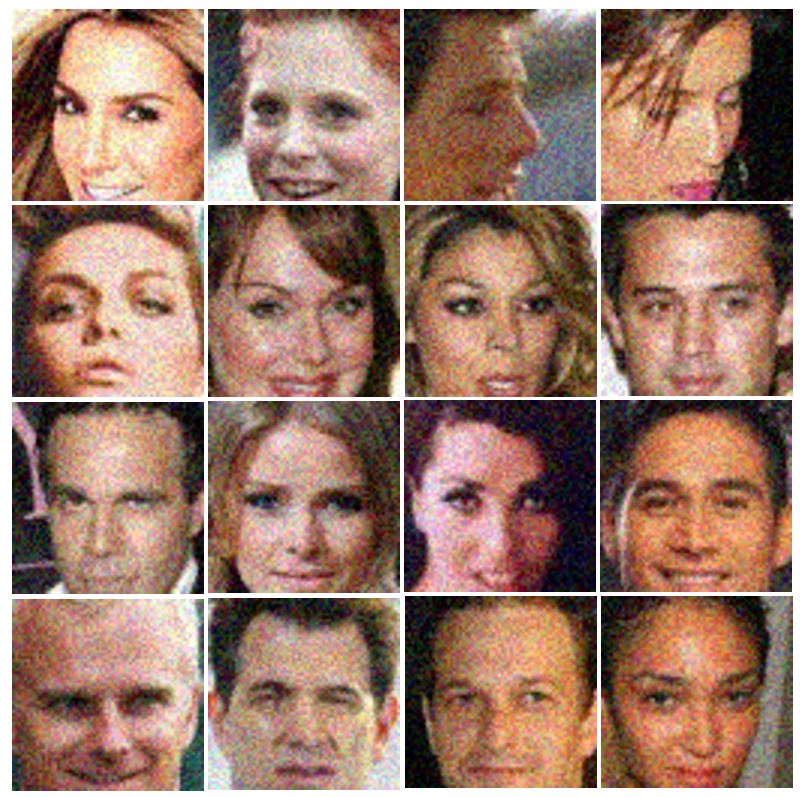}} & \ \ \ & 
			\subfloat[sheltering (FID=74.3)]{\includegraphics[width=1.5in]{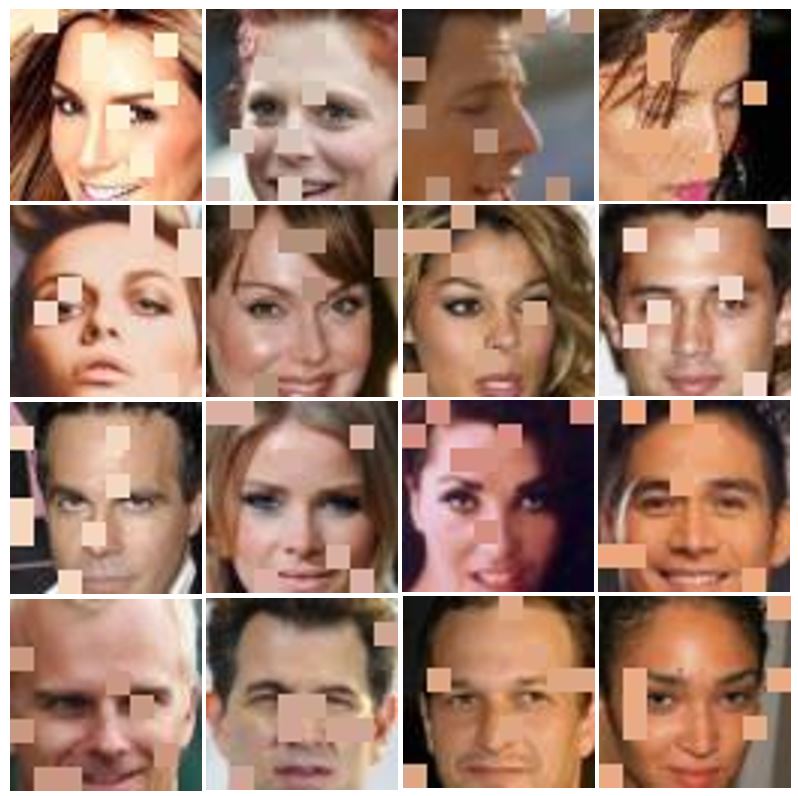}} & \ \ \ &
			\subfloat[exchange (FID=70.9)]{\includegraphics[width=1.5in]{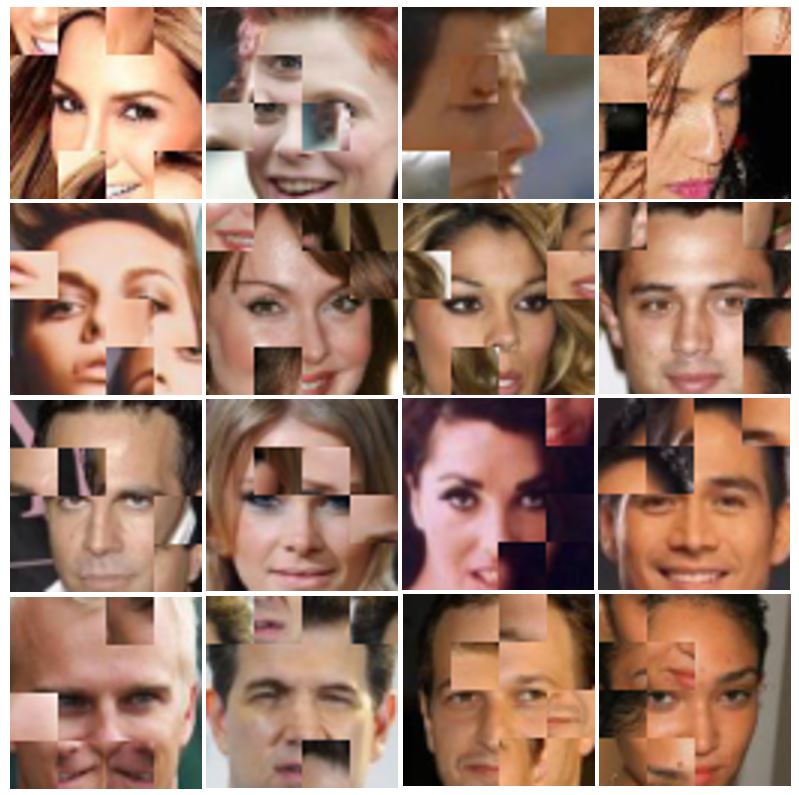}}
		\end{tabular}
		\caption{Examples where FID gives inconsistent results with human judgements ($a<b<c$) on CelebA \cite{liu2015faceattributes}. The ImageNet inception model fails to encode fine-grained features on faces. a) Random noise uniformly distributed in [-33,33] was applied on each pixel. b) Each image was divided into 8x8=64 regions and seven of them were sheltered by a pixel sampled from the face. c) Each image was first divided into 4x4=16 regions and random exchanges were performed twice.}
		\label{fig::hackimagenet}
	\end{figure}

	Results are shown in Fig. \ref{fig::hackimagenet}. With the ImageNet inception model, it is obvious that FID gave inconsistent results with human judgements. In fact, when similar adjustments were conducted with the overall color maintained, FID fluctuated within only a small range. The ImageNet model mainly extracts general features on color, shape to better classify objects in the world while domain-specific facial textures cannot be well represented. 
	
	\label{sec::hackimagenet}
	
	\begin{table}[tb]
		\centering
		\setlength{\abovecaptionskip}{5pt}
		\caption{FID results on different representations. Only the AutoEncoder used in our proposed framework provides consistent results with human judgements.}
		\begin{tabular}{|p{2.5cm}|p{2.0cm}|p{2.0cm}|p{2.0cm}|}
			\hline
			& noise & sheltering & exchange \\
			\hline
			ImageNet & 76 & 74 & \textbf{71} \\
			\hline
			AutoEncoder & \textbf{83} & 21417 & 38609 \\
			\hline
			Discriminator & 122466 & 48322 & \textbf{28557} \\
			\hline
			Human & \textbf{Good} & Bad & Worst \\
			\hline
		\end{tabular}
		\label{tab::fid_rep}
	\end{table}
	
	To attain domain-specific representation, we trained an AutoEncoder on the dataset and used its representation to extract features in our proposed framework. In this experiment, the network architecture of the AutoEncoder is the inverse of the 4-conv DCGAN \cite{DBLP:journals/corr/RadfordMC15} with the feature dimension 1024. For comparison, we also tried to apply the representation of the discriminator after GAN training, which was previously proposed in \cite{che2015mode}. Results are shown in Table \ref{tab::fid_rep}.
	
	It is shown that only representations derived from the AutoEncoder in our proposed framework are effective and give results consistent with human judgements. The discriminator which learns to discriminate fake samples from the real cannot learn good representation for distance measurement. 
	
	To further support our statement that the features encoded by ImageNet model are limited within a low-dimensional manifold, we trained an AutoEncoder with the feature dimension 2048, which is the same as the dimension of features encoded by ImageNet. We again applied the inverse structure of DCGAN \cite{DBLP:journals/corr/RadfordMC15} as the architecture of the AutoEncoder. Principle component analysis (PCA) was conducted on both features encoded by the AutoEncoder and the ImageNet inception model on CelebA \cite{liu2015faceattributes}. Table \ref{tab::pca} shows the percent of explained variance on the first 5 components.
	
	We argue that the ImageNet model should have much greater representation capability than the 4-conv encoder. However, its first two components has relatively higher explained variance (9.35\% and 7.04\%). This supports our claim that the features encoded by ImageNet are limited in a low-dimensional subspace. 
	
	\begin{table}[tb]
		\centering
		\setlength{\abovecaptionskip}{5pt}
		\caption{Results on the explained variance of principle component analysis (PCA) on features encoded by different represenation. Although the architecture of ImageNet model is much more complex than the AutoEncoder, the features encoded by ImageNet model are limited in a relatively low-dimensional subspace. }
		\begin{tabular}{|p{1.7cm}|p{2.0cm}|p{2.0cm}|p{2.0cm}|p{2.0cm}|}
			\hline
			& \multicolumn{2}{c|}{\textbf{AutoEncoder}} & \multicolumn{2}{c|}{\textbf{ImageNet}} \\
			\cline{2-3}\cline{4-5}
			Component & Explained & Accumulated & Explained & Accumulated\\
			\hline
			1 & 5.58\% & 5.58\% & \textbf{9.35\%} & \textbf{9.35\%} \\
			\hline
			2 & 4.66\% & 10.24\% & \textbf{7.04\%} & \textbf{16.39\%} \\
			\hline
			3 & \textbf{3.93\%} & 14.17\% & 3.88\% & \textbf{20.27\%} \\
			\hline
			4 & \textbf{3.66\%} & 17.83\% & 2.67\% & \textbf{22.95\%} \\
			\hline
			5 & \textbf{3.41\%} & 21.24\% & 2.47\% & \textbf{25.42\%} \\
			\hline
		\end{tabular}
		\label{tab::pca}
	\end{table}
	
	Thus, for datasets where images are from a single class such as CelebA \cite{liu2015faceattributes} and LSUN Bedrooms \cite{yu15lsun}, the representation should be acquired via training an AutoEncoder. Our framework employs a domain-specific encoder, which provides more fine-grained information related to specific domain.
	
	\subsection{Study on Evaluation Metric}
	\label{sec::exp_metric}
	In this section, we used the domain-specific representation and studied the improvements of the evaluation metric CAFD proposed in our framework against the state-of-the-art metric FID \cite{NIPS2017_7240}. In datasets with multiple classes, the Gaussian mixture model in CAFD will better fit the feature distribution. Experiments and analysis on both the feature level and the image level were conducted on MNIST. First, we study the distribution of the encoded features via statistical normality test. Then, data is visualized to help get better understanding on the feature space. Finally, A specific case is given where CAFD shows great robustness while FID fails to give consistent results with human judgements.
	
	\subsubsection{Single-manifold vs. Multi-manifold:}
	The Gaussian assumption on the features were commonly used in the literature \cite{DBLP:journals/corr/JinDC15}. Although there are non-linear operations such as relu and max-pooling in the neural network, assuming the normality usually simplifies the model and enables numerical expression. However, in labeled dataset with multiple classes, the single-manifold Gaussian assumption is considered to be over-simplified. 
	
	In this experiment, we performed Anderson-Darling test (AD-test) \cite{AD_test} to quantatively study the normality of the data. Specifically, to test the multivariate normality on a set of features, we performed principle component analysis (PCA) on the data, applied AD-test to the first 10 components and averaged the results. We compared the test results on each class and the whole training set on MNIST. We used a simple 2-conv structure trained on the MNIST classification task as our feature encoder with a output dimension 1024. To reduce the influence of sample numbers on the result, we divided the whole features randomly into 10 sets to study the normality of the mixed features. Results are shown in Table \ref{tab::adtest}. Although the p-value of both features are small, features within a single class get much greater results than the mixed features. It can be inferred that compared to the whole training set, features within each class are much more Gaussian.
	
	\begin{table}[tb]
		\centering
		\setlength{\abovecaptionskip}{5pt}
		\caption{P-value results of AD-test \cite{AD_test} on features of each class and the whole training images. The whole features were randomly divided into 10 sets. Compared to the mixed features, features encoding images from a single class are more like a single-manifold Gaussian structure.}
		\begin{tabular}{*{6}{c}}
			\hline
			Set Number & 0 & 1 & 2 & 3 & 4\\
			\hline
			Class & $5.0\times 10^{-2}$ & $2.3 \times 10^{-10}$ & $8.5 \times 10^{-4}$ & $5.1\times 10^{-2}$ & $2.6\times 10^{-2}$ \\
			\hline
			Mixed & $6.0\times 10^{-11}$ & $3.8\times 10^{-13}$ & $1.2\times 10^{-18}$ & $1.7\times 10^{-14}$ & $1.3\times 10^{-14}$ \\
			\hline
			Set Number & 5 & 6 & 7 & 8 & 9 \\
			\hline
			Class & $7.0\times 10^{-2}$ & $6.3\times 10^{-3}$ & $6.1\times 10^{-4}$ & $3.0\times 10^{-3}$ & $4.6\times 10^{-4}$ \\
			\hline
			Mixed & $2.2\times 10^{-11}$ & $3.9\times 10^{-13}$ & $1.8\times 10^{-16}$ & $6.1\times 10^{-11}$ & $2.1\times 10^{-14}$ \\
			\hline
		\end{tabular}
		\label{tab::adtest}
	\end{table}

\begin{table}[tb]
	\centering
	\caption{P-value results of mardia test \cite{mardia1985mardia} on features of each class and the whole test images. The whole features were randomly divided into 10 sets.}
	\begin{tabular}{*{6}{c}}
		\hline
		Set Number & 0 & 1 & 2 & 3 & 4\\
		\hline
		Class & $1.2\times 10^{-94}$ & $<10^{-300}$ & $8.0 \times 10^{-38}$ & $8.3\times 10^{-46}$ & $6.5\times 10^{-78}$ \\
		\hline
		Mixed & $7.7\times 10^{-231}$ & $2.2\times 10^{-208}$ & $4.6\times 10^{-214}$ & $2.2\times 10^{-209}$ & $7.5\times 10^{-235}$ \\
		\hline
		Set Number & 5 & 6 & 7 & 8 & 9 \\
		\hline
		Class & $3.1\times 10^{-47}$ & $9.3\times 10^{-68}$ & $2.2\times 10^{-89}$ & $2.7\times 10^{-63}$ & $4.4\times 10^{-105}$ \\
		\hline
		Mixed & $1.0\times 10^{-290}$ & $8.8\times 10^{-246}$ & $2.7\times 10^{-251}$ & $3.2\times 10^{-267}$ & $2.0\times 10^{-217}$ \\
		\hline
	\end{tabular}
	\label{tab::mardiatest}
\end{table}
	
	In addition, we used mardia test \cite{mardia1985mardia} in the R package MVN \cite{korkmaz2014mvn} to directly study the multivariate normality. We first performed principle component analysis (PCA) on both images within a class and the whole test set respectively. Then, mardia test \cite{mardia1985mardia} was used to assess the multivariate normality of the first 5 components. Results (shown in Table \ref{tab::mardiatest}) are consistent with previous experiments on AD-test \cite{AD_test}. Both normality tests suggested that compared to a single-manifold multivariate Gaussian model, the overall features are better fitted with a multi-manifold Gaussian mixture model. Thus, the basic assumption of CAFD in our framework is more reasonable than the FID \cite{NIPS2017_7240} method. 
	
	\subsubsection{Feature Visualization:}
\begin{figure}[tb]
		\centering
		\includegraphics[width=3.5in]{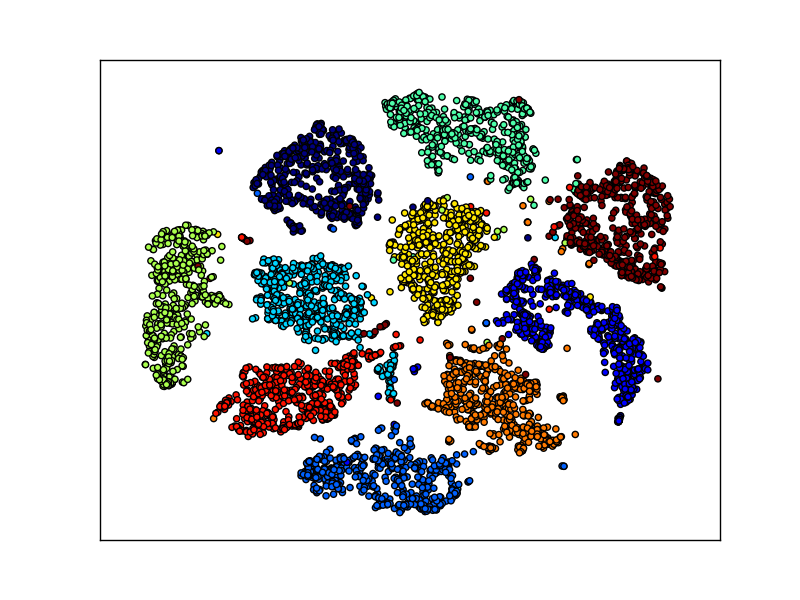}
		\caption{Visualization of the features encoding the training set on MNIST via t-sne \cite{maaten2008visualizing}. Features are distributed in groups by their class labels.}
		\label{fig::visual}
	\end{figure}

	To get intuitive understanding towards the multi-manifold structure on the feature distribution, we performed feature visualization via t-sne \cite{maaten2008visualizing} on MNIST training set and colored them by their class labels. As shown in Fig. \ref{fig::visual}, it is clear that features encoding images from the same class cluster together and the whole features are more like a mixture of ten independent distribtuions with their own class centers. Therefore, assuming the normality on the whole features is considered to be over-simplified. The encoder tends to cluster features from the same class and the overall distribution is multi-manifold in a group manner.
	
	\begin{figure}[tb]
		\centering
		\begin{tabular}{ccc}
			\subfloat[generated (FID=73.11)]{\includegraphics[width=1.5in]{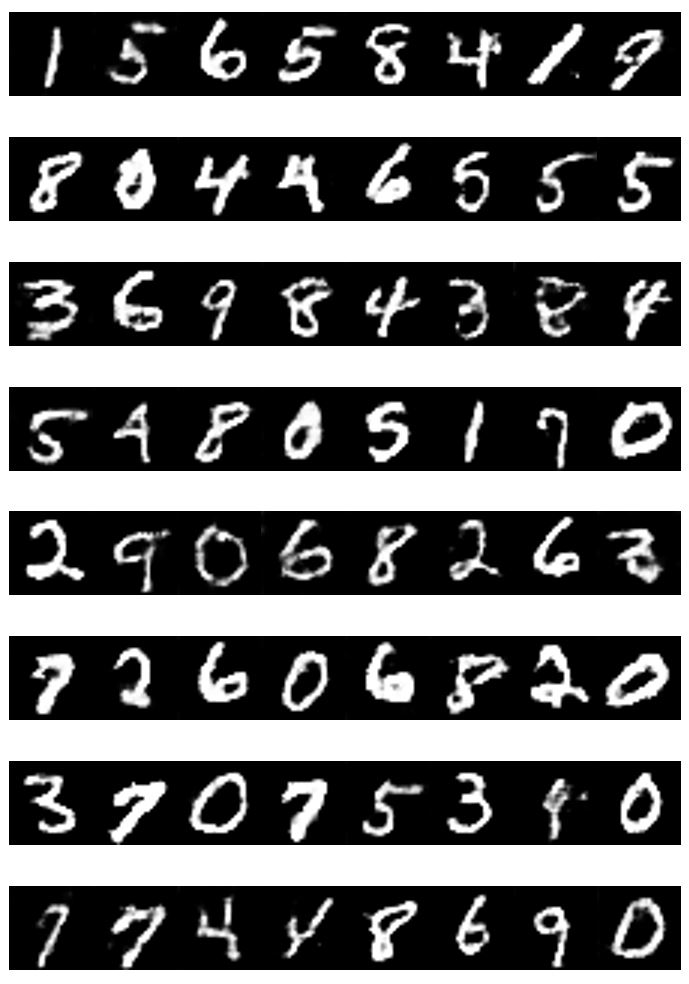}} & \ \ \ \ \ \ \ \ \ &
			\subfloat[hack (FID=72.82)]{\includegraphics[width=1.5in]{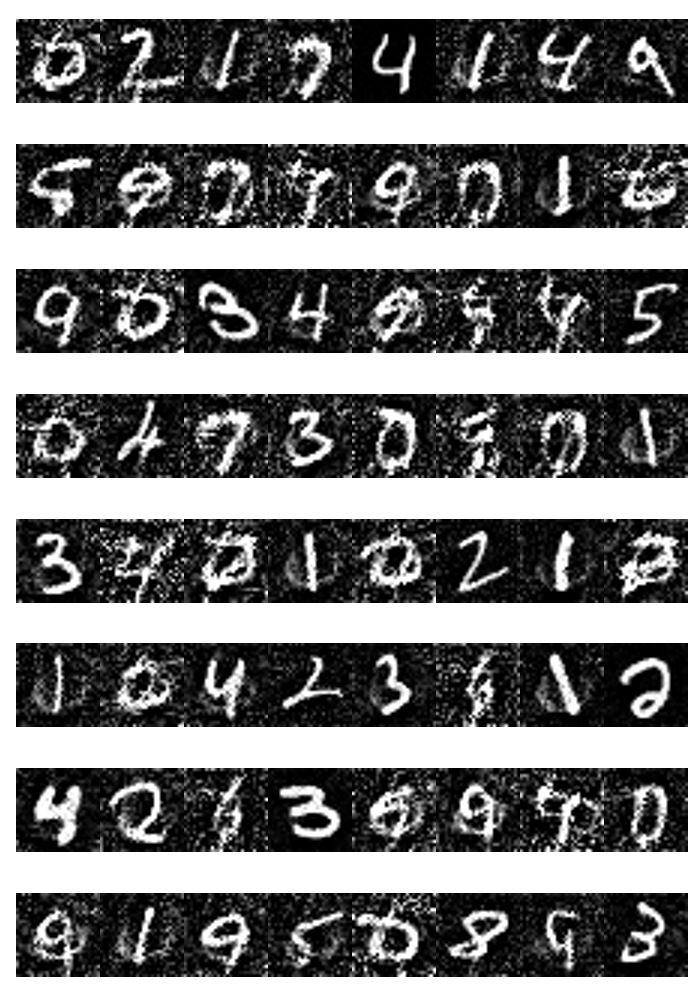}}
		\end{tabular}
		\setlength{\belowcaptionskip}{2pt}
		\caption{Examples where FID gives inconsistent results with human judgements on MNIST. Due to the over-simplified Gaussian assumption, FID can be hacked by mode collapse. a) Samples generated by a DCGAN model. b) Handmade images via axis permutation and FGSM \cite{goodfellow2014explaining}. }
		\label{fig::hackfid}
	\end{figure}
	
	\subsubsection{Comparison between FID and CAFD:}
	\label{sec::hackfid}
	
	In this experiment, we designed cases where FID fails to give consistent results with human judgements. FID, as a overall statistical measure, is able to detect either a single mode dropping or a trivial linear combination of two images. However, as its formulation has relatively limited constraints, it may be hacked by complicated situations. 
	
	Considering the features extracted from MNIST test data, which has a zero FID with itself. We performed operations below on the features.
	
	\begin{enumerate}[Step 1]
		\item Performed principle component analysis (PCA) on the original features.
		\item Normalized each axis to zero mean and unit variance.
		\item Switched the normalized projection of the first two component.
		\item Unnormalized the data and reconstructed features.
	\end{enumerate}
	
	The adjusted features are completely different with the original one with zero FID maintained. The over-simplified Gaussian assumption on overall distribution cannot tell the differences while our proposed method is able to report the changes with CAFD raising from 0 to 539.8 (See Table \ref{tab::cafd}). 
	
	We used FGSM \cite{goodfellow2014explaining} to reconstruct the images from the adjusted features. Specifically, we first trained an decoder for initialization via an AutoEncoder with the encoder fixed. Then, we performed pixelwise adjustment via FGSM \cite{goodfellow2014explaining} to lower the reconstruction error. Because the used encoder has a relatively simple structure, the final reconstruction error is still relatively high after optimized. We trained a simple DCGAN \cite{DBLP:journals/corr/RadfordMC15} model and took samples (generated by intermediate models during training) with comparable FID with our constructed images. Results are shown in Fig. \ref{fig::hackfid}.
	
	\begin{table}[tb]
		\centering
		\setlength{\abovecaptionskip}{5pt}
		\caption{Results of FID, CAFD and KLD on MNIST. Lower scores infer better image quality. The `test' denotes the MNIST test set, `adjusted' denotes the features after axis permutation. `generated' and `hack' are the sampled images in Fig. \ref{fig::hackfid}. Compared to FID, CAFD are more robust to feature-level adjustments.}
		\begin{tabular}{|p{2.5cm}|p{2.5cm}|p{2.5cm}|p{2.5cm}|}
			\hline
			& FID & CAFD & $KL(p(y^*)||p(y))$ \\
			\hline
			test & 0 & 0 & 0 \\
			\hline
			adjusted & 0 & 539.8 & 0.03675 \\
			\hline
			generated & 73.1 & 201.4 & 0.001893 \\
			\hline
			hack & 72.8 & 468.6 & 0.04941 \\
			\hline
			train & 22.0 & 99.8 & 0.000572 \\
			\hline
		\end{tabular}
		\label{tab::cafd}
	\end{table}
	
	It is obvious that the quality of constructed images are much worse than the generated samples. After axis permutation, the constructed images suffers from mode collapse. There are many pictures in the right which resemble more than one digits and are hard to recognize. However, it still received a FID of 72.82 lower than that (73.11) received by generated samples. CAFD and KLD results on these cases are shown in Table \ref{tab::cafd}. While FID gives misleading results, CAFD are much more robust on the adjusted features. Compared to the constructed images (468.6), the generated images received a much lower CAFD (201.4), which is consistent with human judgements. This demonstrates the improved effectiveness of the evaluation metric in our proposed framework. 
	
	\section{Conclusions}
	In this paper, we have presented an improved evaluation framework for Generative Adversarial Networks, which improves conventional methods on both representation and evaluation metric. We argue that a domain-specific encoder is needed and propose Class-Aware Frechet Distance to better fit the feature distribution. To our best knowledge, we are the first to provide counter examples where the state-of-the-art FID is inconsistent with human judgements. Experiments and analysis on both the feature level and the image level have shown that our framework is more effective than FID and improves its robustness. 

	\clearpage

\bibliographystyle{splncs}
\bibliography{egbib}

\section*{Appendix}
\appendix
		
\section{A Benchmark for Popular GANs}
\begin{figure}[tb]
	\centering
	\begin{tabular}{cc}
		\subfloat[Results on MNIST]{\includegraphics[width=2.4in]{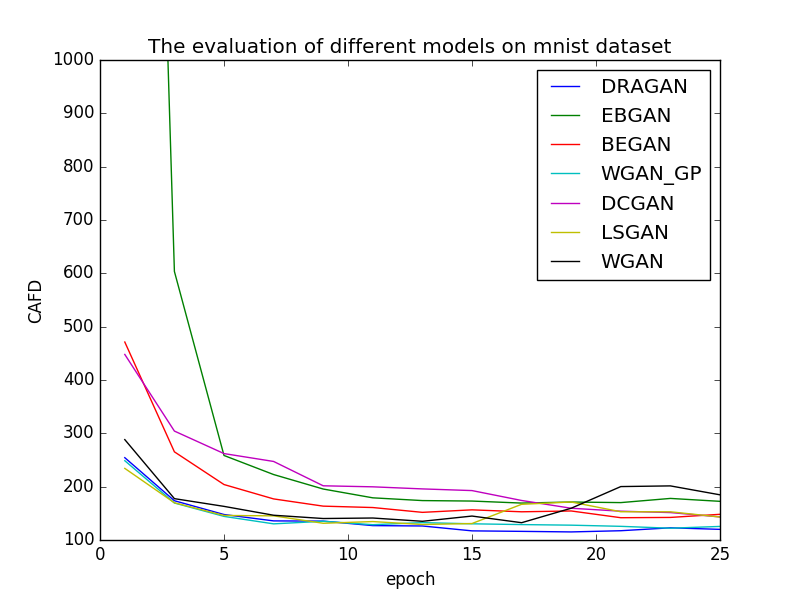}} & 
		\subfloat[Results on FASHION-MNIST \cite{xiao2017/online}]{\includegraphics[width=2.4in]{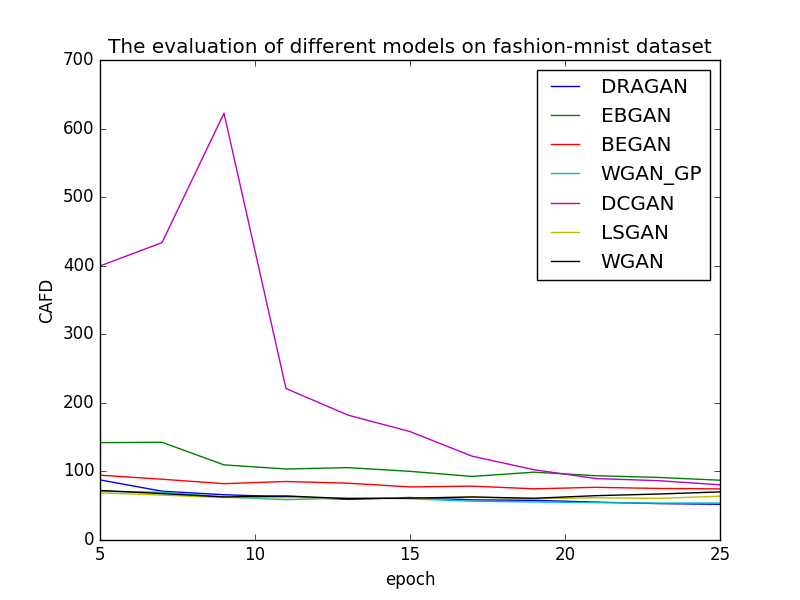}}
	\end{tabular}
	\caption{Results of our evaluation framework on popular GAN models. The experiments were performed on MNIST and FASHION-MNIST \cite{xiao2017/online}. }
	\label{fig::res}
\end{figure}

\begin{table}[tb]
	\centering
	\setlength{\abovecaptionskip}{5pt}
	\caption{CAFD Results of different GAN models on MNIST and FASHION-MNIST \cite{xiao2017/online}. The encoder was specifically trained on the dataset.}
	\begin{tabular}{p{3.0cm}p{3.5cm}p{3.5cm}}
		\hline
		& MNIST & FASHION-MNIST \cite{xiao2017/online} \\
		\hline
		DCGAN \cite{DBLP:journals/corr/RadfordMC15} & $143.7\pm 1.6$ & $80.3\pm 0.4$ \\
		\hline
		LSGAN \cite{Mao_2017_ICCV} & $143.4\pm 0.6$ & $64.1\pm 0.4$ \\
		\hline
		BEGAN \cite{DBLP:journals/corr/BerthelotSM17} & $147.1\pm 2.1$ & $75.0\pm 0.4$ \\
		\hline
		EBGAN \cite{zhao2016energy} & $172.7\pm 2.4$ & $86.4\pm 0.5$ \\
		\hline
		DRAGAN \cite{DBLP:journals/corr/KodaliAHK17} & \textbf{120.2$\pm$0.6} & \textbf{51.9$\pm$0.4} \\
		\hline
		WGAN \cite{arjovsky2017wasserstein} & $184.5\pm 0.8$ & $69.5 \pm 0.5$ \\
		\hline
		WGAN-GP \cite{NIPS2017_7159} & \textbf{126.7$\pm$0.9} & \textbf{54.1$\pm$0.5} \\
		\hline
	\end{tabular}
	\label{tab::res}
\end{table}

In order to benmark the performance of GANs on generating domain-specific images, we conducted experiments on 7 popular GAN models\footnote{We used the off-the-shelf tensorflow package \url{https://github.com/hwalsuklee/tensorflow-generative-model-collections}.} including DCGAN \cite{DBLP:journals/corr/RadfordMC15}, LSGAN \cite{Mao_2017_ICCV}, BEGAN \cite{DBLP:journals/corr/BerthelotSM17}, EBGAN \cite{zhao2016energy}, DRAGAN \cite{DBLP:journals/corr/KodaliAHK17}, WGAN \cite{arjovsky2017wasserstein}, WGAN-GP \cite{NIPS2017_7159}. Our experiments were performed on MNIST and FASHION-MNIST \cite{xiao2017/online}. We will include other popular datasets such as CIFAR-10 \cite{krizhevsky2009learning}, CelebA \cite{liu2015faceattributes} and ImageNet \cite{ILSVRC15} in the future.

Results are shown in Fig. \ref{fig::res} and Table \ref{tab::res}. All of the tested models converge well. DCGAN \cite{DBLP:journals/corr/RadfordMC15}, which is the first to introduce convolutional neural networks into generative models, struggles more on convergence than the newly proposed GAN variants. DRAGAN \cite{DBLP:journals/corr/KodaliAHK17} and WGAN-GP \cite{NIPS2017_7159} get the top two scores on both datasets. Both BEGAN \cite{DBLP:journals/corr/BerthelotSM17} and WGAN \cite{arjovsky2017wasserstein} focus more on stable training, while the qualities of their generated images are not the best. WGAN-GP \cite{NIPS2017_7159} improves WGAN \cite{arjovsky2017wasserstein} by using norm penalizing to replace weight clipping. It generates higher quality images compared to its baseline. DRAGAN \cite{DBLP:journals/corr/KodaliAHK17} utilizes a gradient penalty scheme and mitigates the problem of mode collapse. It is worth noting that the recently proposed DRAGAN \cite{DBLP:journals/corr/KodaliAHK17} and WGAN-GP \cite{NIPS2017_7159} outperform other models by a relatively large margin. We can infer that the development of exploring better GAN architectures and training strategies is still highly active.

\section{Qualitative Visualization}
\begin{figure}
	\centering
	\begin{tabular}{cc}
		\subfloat[$CAFD=0.0$]{\includegraphics[width=2.2in]{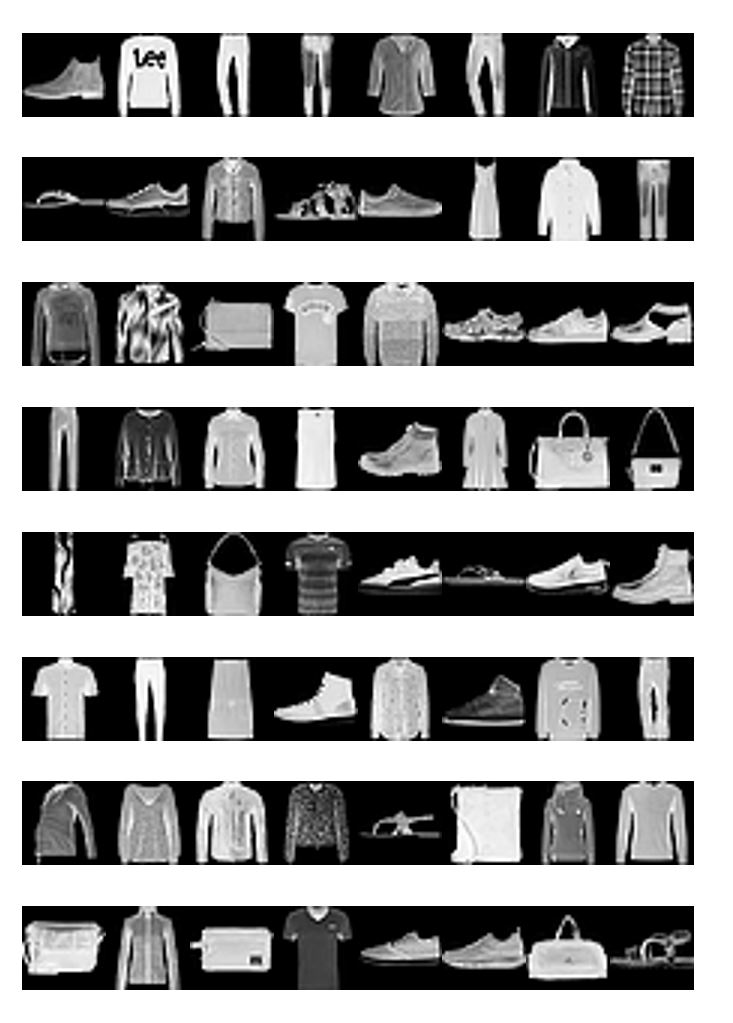}} & 
		\subfloat[$CAFD=52.10$]{\includegraphics[width=2.2in]{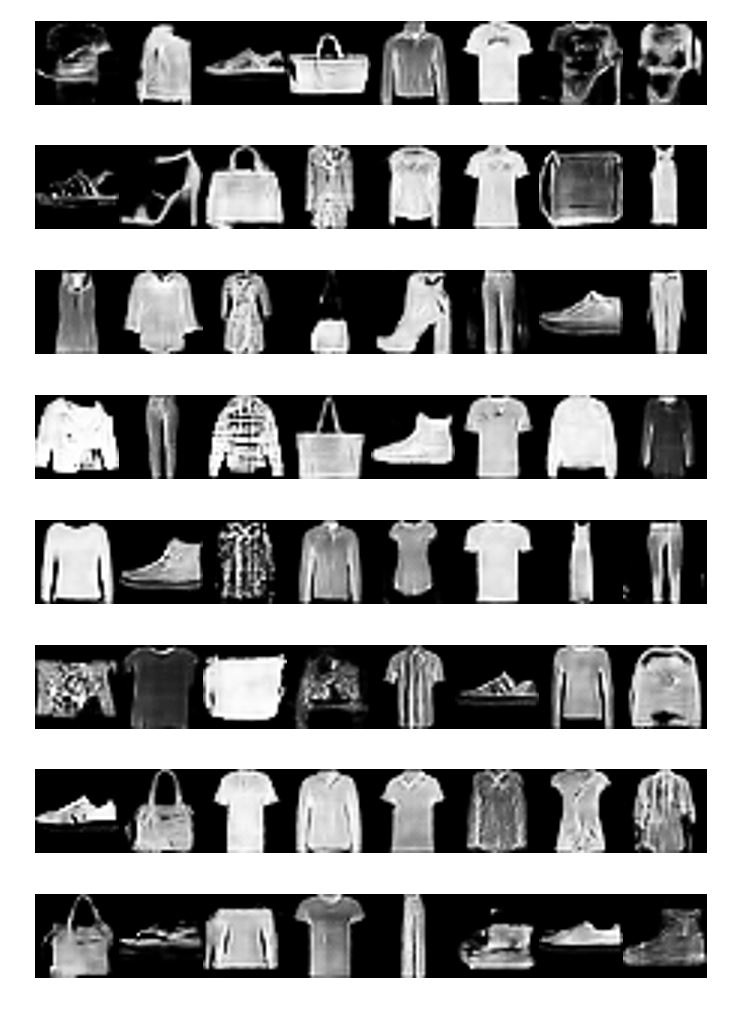}}
	\end{tabular}
	\caption{Qualitative visualization of different scores on FASHION-MNIST \cite{xiao2017/online}.}
	\label{fig::fashion_1}
\end{figure}

\begin{figure}[tb]
	\centering
	\begin{tabular}{cc}
		\subfloat[$CAFD=100.14$]{\includegraphics[width=2.2in]{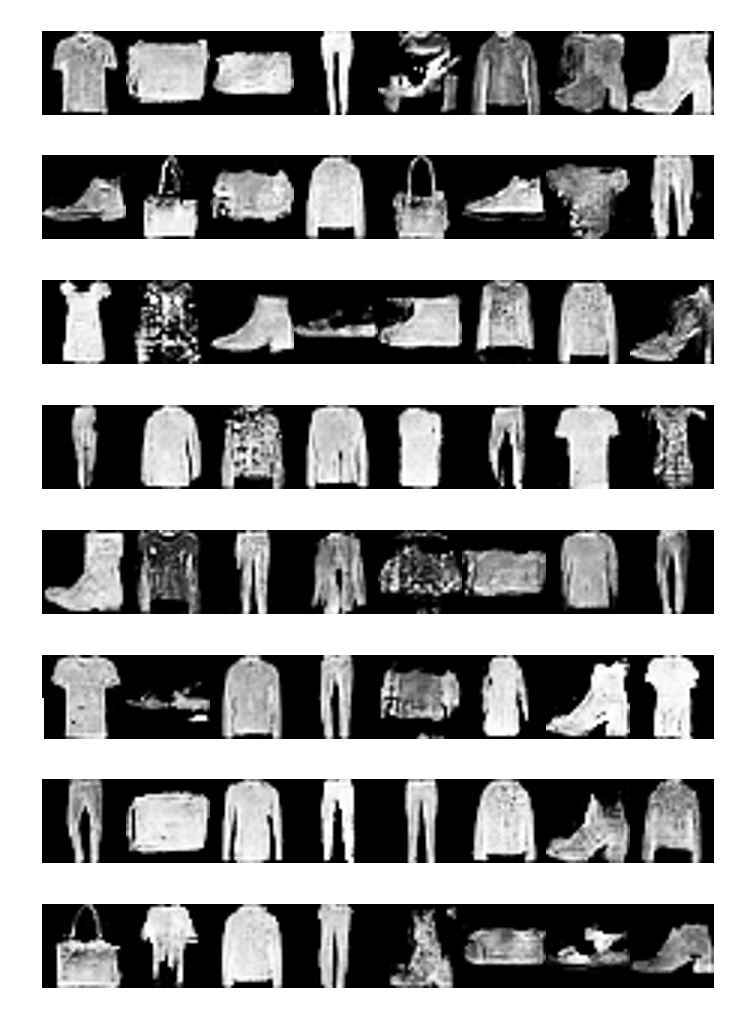}} & 
		\subfloat[$CAFD=151.33$]{\includegraphics[width=2.2in]{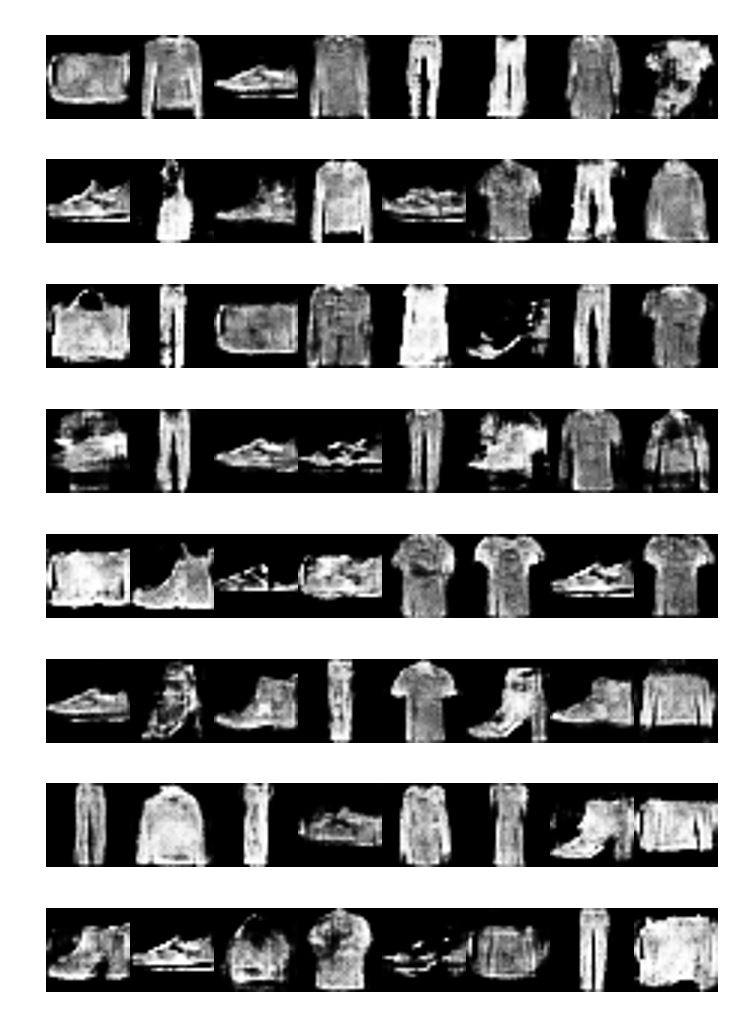}} \\
		\subfloat[$CAFD=188.51$]{\includegraphics[width=2.2in]{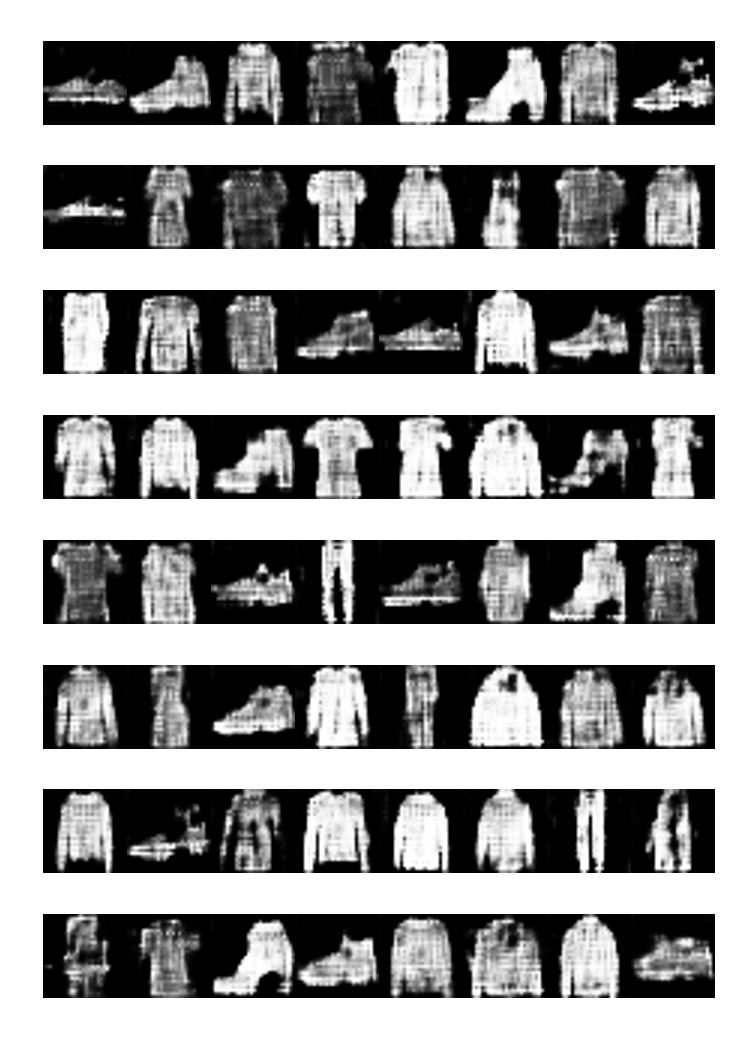}} & 
		\subfloat[$CAFD=220.89$]{\includegraphics[width=2.2in]{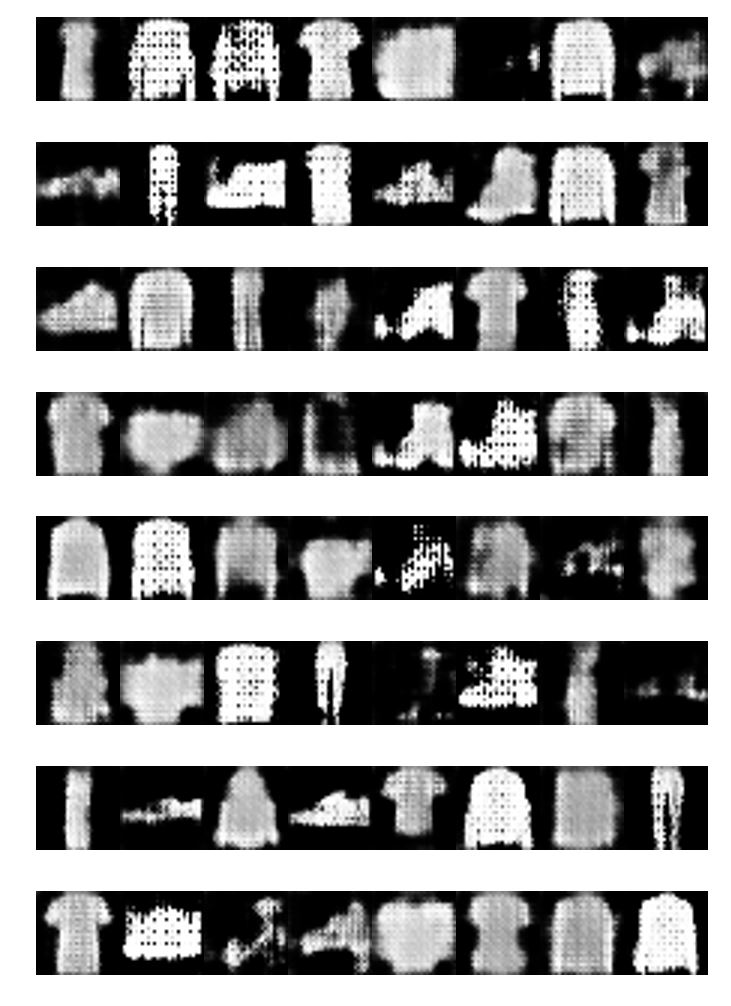}}
	\end{tabular}
	\caption{Qualitative visualization of different scores on FASHION-MNIST \cite{xiao2017/online}.}
	\label{fig::fashion_2}
\end{figure}

\begin{figure}[tb]
	\centering
	\begin{tabular}{cc}
		\subfloat[$CAFD=276.73$]{\includegraphics[width=2.2in]{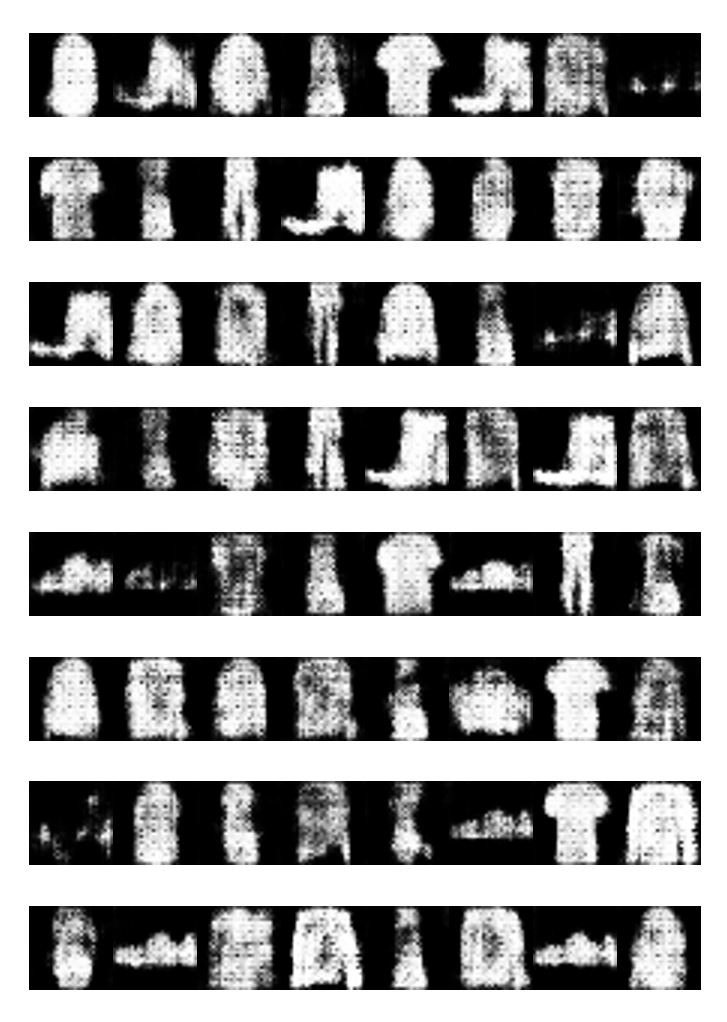}} & 
		\subfloat[$CAFD=305.83$]{\includegraphics[width=2.2in]{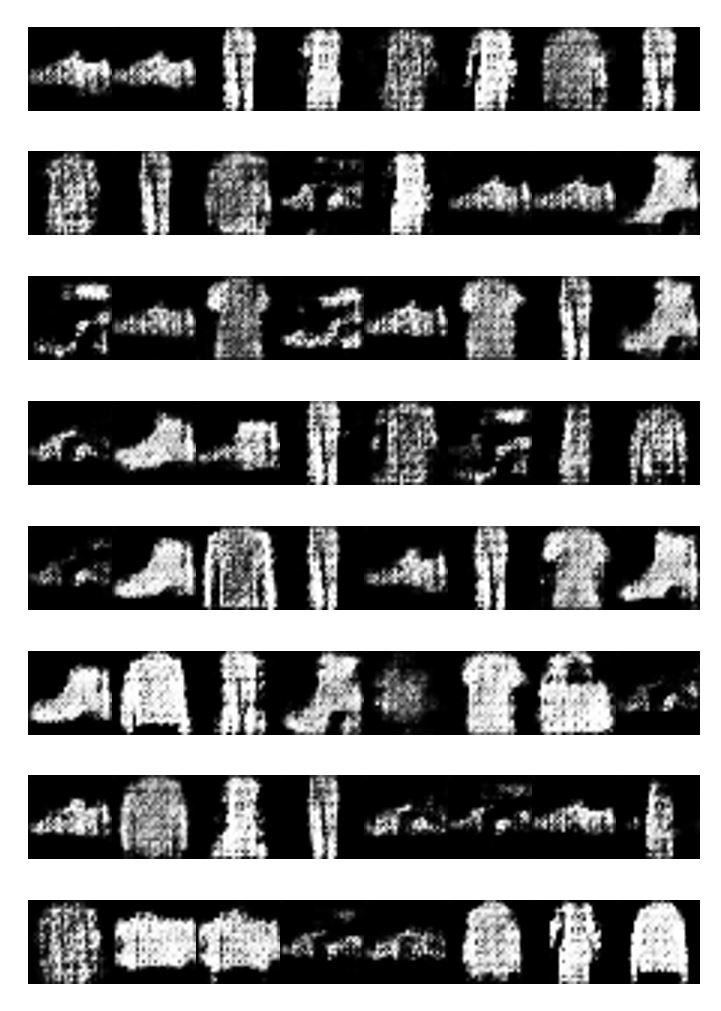}} \\
		\subfloat[$CAFD=399.65$]{\includegraphics[width=2.2in]{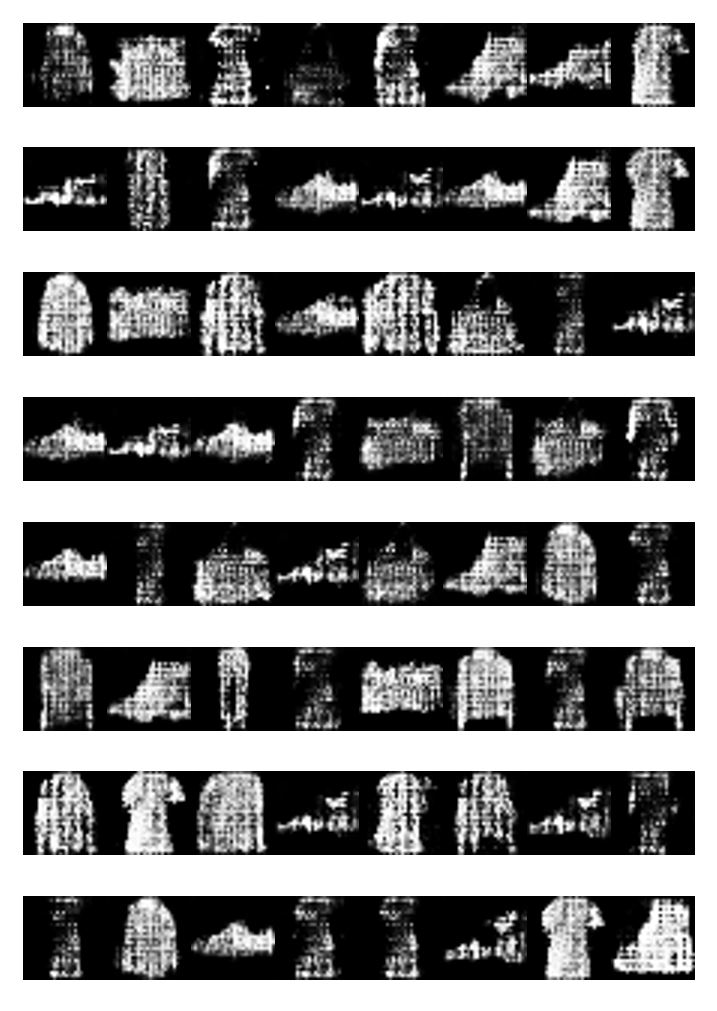}} & 
		\subfloat[$CAFD=622.40$]{\includegraphics[width=2.2in]{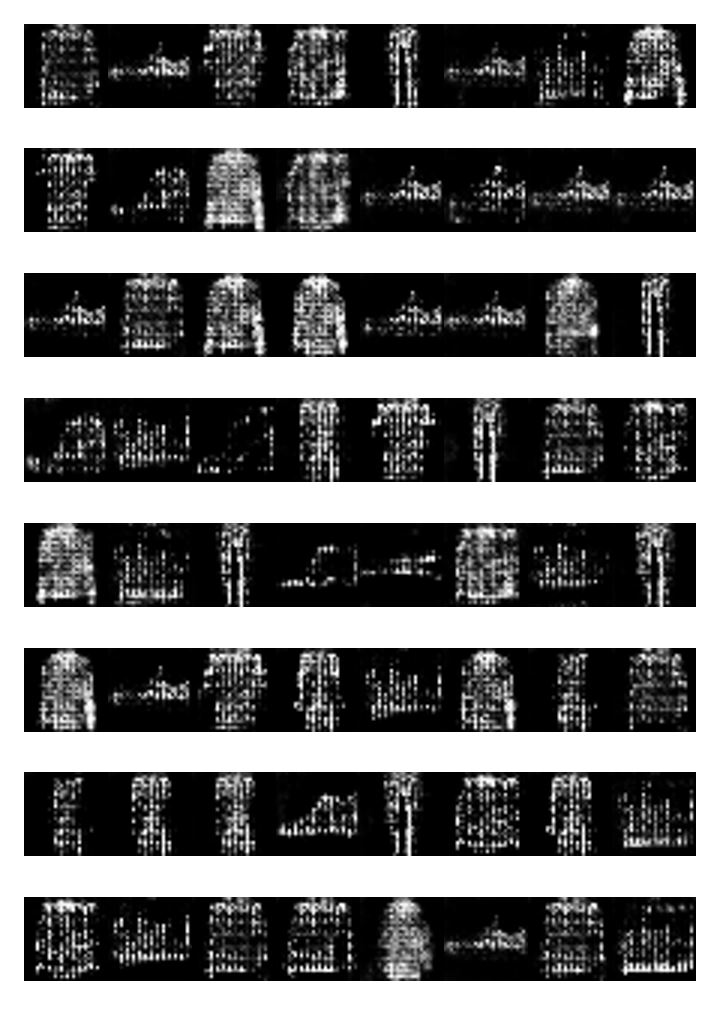}}
	\end{tabular}
	\caption{Qualitative visualization of different scores on FASHION-MNIST \cite{xiao2017/online}.}
	\label{fig::fashion_3}
\end{figure}

In this section, we provide qualitative visualization of images with different scores under our evaluation framework. Images were generated by intermediate models during GAN training. Experiments were conducted on FASHION-MNIST \cite{xiao2017/online}. Figs. \ref{fig::fashion_1}, \ref{fig::fashion_2} and \ref{fig::fashion_3} show the results. The Class-Aware Frechet Distance (CAFD) metric in our proposed framework gives consistent results with human judgements.

\clearpage

\end{document}